\documentclass[letterpaper]{article} 
\usepackage{aaai24}  
\usepackage{times}  
\usepackage{helvet}  
\usepackage{courier}  
\usepackage[hyphens]{url}  
\usepackage{graphicx} 
\usepackage{natbib}  
\usepackage{caption} 
\usepackage{algorithm}
\usepackage{algorithmic}
\usepackage{makecell}
\usepackage{newfloat}
\usepackage{listings}
\DeclareCaptionStyle{ruled}{labelfont=normalfont,labelsep=colon,strut=off} 
\floatstyle{ruled}
\newfloat{listing}{tb}{lst}{}
\floatname{listing}{Listing}
\usepackage{booktabs}
\usepackage{arydshln}
\usepackage{bbding}
\usepackage{multirow}
\usepackage{pifont}

\newcommand{\tabincell}[2]{\begin{tabular}{@{}#1@{}}#2\end{tabular}}
\newcommand{\cmark}{\ding{51}}
\newcommand{\xmark}{\ding{55}}
\usepackage{amsmath,amsfonts}

\title{DHGCN: Dynamic Hop Graph Convolution Network for \\Self-Supervised Point Cloud Learning}
\author{
    Jincen Jiang\textsuperscript{\rm 1}\equalcontrib,
    Lizhi Zhao\textsuperscript{\rm 1}\equalcontrib,
    Xuequan Lu\textsuperscript{\rm 2}\thanks{Corresponding authors.},
    Wei Hu\textsuperscript{\rm 3},
    Imran Razzak\textsuperscript{\rm 4},
    Meili Wang\textsuperscript{\rm 1}\footnotemark[2]
}
\affiliations{
    \textsuperscript{\rm 1}Northwest A\&F University,\\
    \textsuperscript{\rm 2}La Trobe University,\\ 
    \textsuperscript{\rm 3}Peking University,\\
    \textsuperscript{\rm 4}University of New South Wales\\
    \{jinec, zhaolizhi, wml\}@nwsuaf.edu.cn, b.lu@latrobe.edu.au,
    forhuwei@pku.edu.cn, 
    imran.razzak@unsw.edu.au
}

\begin{document}

\maketitle

\begin{abstract}
Recent works attempt to extend Graph Convolution Networks (GCNs) to point clouds for classification and segmentation tasks. These works tend to sample and group points to create smaller point sets locally and mainly focus on extracting local features through GCNs, while ignoring the relationship between point sets. In this paper, we propose the Dynamic Hop Graph Convolution Network (DHGCN) for explicitly learning the contextual relationships between the voxelized point parts, which are treated as graph nodes. Motivated by the intuition that the contextual information between point parts lies in the pairwise adjacent relationship, which can be depicted by the hop distance of the graph quantitatively, we devise a novel self-supervised part-level hop distance reconstruction task and design a novel loss function accordingly to facilitate training. In addition, we propose the Hop Graph Attention (HGA), which takes the learned hop distance as input for producing attention weights to allow edge features to contribute distinctively in aggregation. Eventually, the proposed DHGCN is a plug-and-play module that is compatible with point-based backbone networks. Comprehensive experiments on different backbones and tasks demonstrate that our self-supervised method achieves state-of-the-art performance. \textit{Our source code is available at}: https://github.com/Jinec98/DHGCN. 
\end{abstract}


\begin{figure}[!h]
\centering
\begin{minipage}[b]{0.9\linewidth}
\begin{center}
\includegraphics[width=0.85\linewidth]{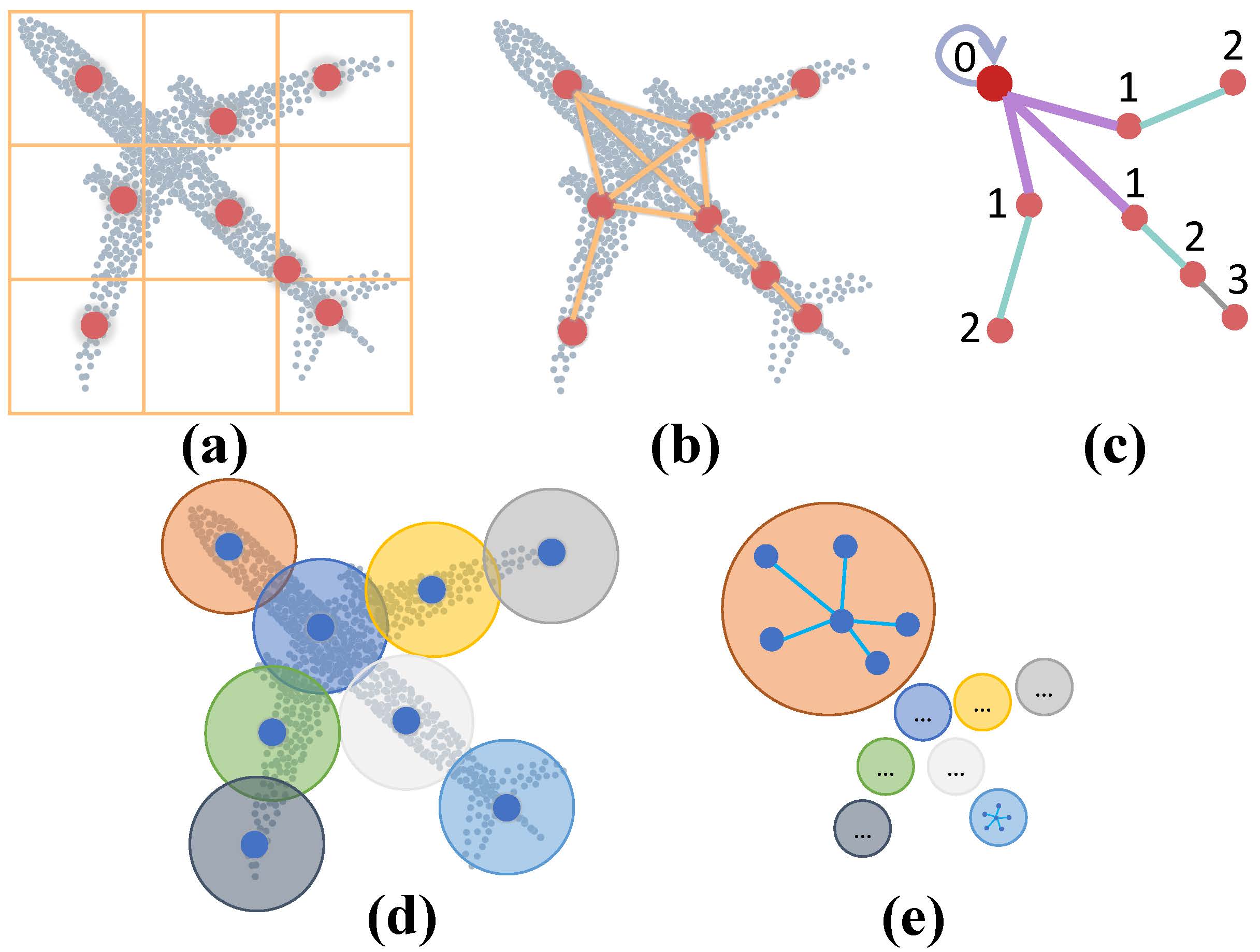}
\end{center}
\end{minipage}
\caption{
First row: Constructing the ground truth graph for our self-supervised hop distance reconstruction task. (a): Voxelizing the point cloud into parts, taking each part as a graph node. (b): The topology of the ground truth graph. Two nodes are adjacent if their scaled bounding boxes are intersected. (c): The shortest path between a node (enlarged red point) and other nodes.
The number on each node denotes the hop distance which motivates our self-supervised task. 
Second row: Sampling and grouping based strategy \cite{dgcnn, wang2019graphattenconv}. (d): Sampling center points and grouping local point sets. (e): Constructing a local graph for each point set. 
We explore the contextual relationships between parts, while previous strategies focus on extracting local features of point sets.
}
\label{fig:intro_pic}
\end{figure}

\section{Introduction}
\label{sec:intro}
A point cloud is an unordered collection of scattered points representing geometric information in the 3D space. 
Point cloud processing and understanding are crucial in many areas, such as autonomous driving, virtual reality, etc.
Unlike regular 2D pixels on the images, point clouds typically have irregular point distributions, making it difficult to directly apply traditional Convolution Neural Networks (CNNs) to point clouds for extracting features \cite{shi2020pointgnn,de2023iterativepfn,zhang2020pointfilter}.

In addition, the complex geometric structures of the point cloud can be well-structured with graphs by encoding the representations of pairwise relationships of points  \cite{defferrard2016convolutional,chung1997spectral}. 
Therefore, researchers have made efforts to generalize Graph Convolution Networks (GCNs) to point clouds for classification and segmentation, achieving encouraging results  \cite{qi20173d,landrieu2018large,bi2019graph,yi2023graph}.
DGCNN \cite{dgcnn} constructs a \(k\)-Nearest Neighbor (\(k\)NN) graph for each center point, and 
captures graph structure by propagating and aggregating the offset relationships between a center point and its neighbors. 
\citet{wang2019graphattenconv} assigns attention weights to different neighboring points and feature channels. 
These works tend to sample and group points to create point sets locally, and construct a graph for each set (Figure \ref{fig:intro_pic} (d)-(e)). 
They usually focus on extracting graph representations for local point sets through GCNs, while ignoring the explicit contextual relationships between them.

In this paper, we attempt to embed the distance in geometric space between point sets explicitly to learn their contextual relationships (e.g., topology, adjacency, etc).
Our intuition is that the geometric distance of point sets lies in the pairwise adjacent relationship between them.
In addition, with considering point sets as nodes and representing the point cloud as a graph, the hop distance concept in graph theory can depict 
the degree of adjacency quantitatively.

Motivated by this intuition, we propose a novel \textit{self-supervised part-level hop distance reconstruction task}.
We design the Dynamic Hop Graph Convolution Network (DHGCN) for extracting and embedding the low-level geometric distance to learn the high-level contextual relationships between voxelized point parts. 
In the pre-processing stage, we first split the entire point cloud into voxel parts, and construct a \textit{ground truth graph} with each part serving as a graph node to compute the distance matrix, which can imply the degree of adjacency between two nodes quantitatively (Figure \ref{fig:intro_pic} (a)-(c)). 
Next, given an input point cloud, we construct a complete graph with randomly initialized distance matrix (i.e., no contextual information) 
and attempt to predict the hop distance matrix to learn the part-level contextual relationship.
We devise a hop distance loss to supervise the predicted distance matrix that is dynamically updated in each layer. 
Further, we design Hop Graph Attention (HGA) that takes the learned distance matrix as input for assigning more attention to edge features between neighboring parts (i.e., parts with short distances) and less attention to distant parts,  allowing edge features to contribute distinctively in aggregation.
Finally, we make our DHGCN compatible with point-based backbone networks through pooling and repeating, making it a plug-and-play module.

The main contributions of this paper are as follows. 
\begin{itemize}
\item We propose a novel self-supervised hop distance reconstruction task and a hop distance loss for learning the contextual relationships between point parts, by considering the hop distance as the proxy to depict the degree of adjacency quantitatively. 
\item We propose Hop Graph Attention, a module that takes the dynamically learned hop distance as input to produce attention weights, allowing edge features to contribute distinctively in aggregation. 
\item We make our DHGCN a plug-and-play module that can be easily embedded in point-based backbone networks. 
Extensive experiments show that our self-supervised DHGCN achieves state-of-the-art performance on different downstream tasks. 
\end{itemize}

\section{Related Work}

\subsection{Self-supervised Point Cloud Learning}
Point-based methods are pioneered by PointNet \cite{qi2017pointnet}, which consumes raw point cloud as input using shared multi-layer perceptions (MLPs). 
PointNet++ \cite{pointnet++} further devises a hierarchical architecture that recursively samples point sets to capture multi-scale local geometric information.
As the cornerstone of point-based methods, PointNet++ has inspired numerous modern works.

Self-supervised learning (SSL) methods for point clouds aim to learn point cloud intrinsic representations through well-designed pretext tasks without labeled data.
Recent works can be roughly summarized as contrastive and reconstructive methods \cite{wu2021self}.
The contrastive methods contrast the latent representations of different point cloud transformation views (e.g., rotation, jitter, scale, etc.) and design pretext tasks based on inter-data information such as similarity \cite{xie2020pointcontrast, chen20214dcontrast,gao2020graphter}. 
CrossPoint \cite{afham2022crosspoint} enforces the correspondence between a point cloud and its rendered 2D image while preserving the model's invariance to spatial transformations.
The reconstructive methods typically aim to reconstruct intra-data information from low-quality (e.g., mask, noise, etc.) input, which exploits the point cloud intrinsic structure as self-supervised signals \cite{yang2018foldingnet, yu2022point, pang2022masked}. 
OcCo \cite{wang2021unsupervised} generates occluded point clouds from randomly sampled camera views and trains an encoder-decoder model to complete the original point clouds.

\subsection{Graph-based Learning Methods}
Graphs are the universal representations of heterogeneous pairwise relationships of non-Euclidean data, such as point clouds \cite{defferrard2016convolutional}. Previous works extend convolution from 2D CNNs to graphs by processing the graph spectral representation \cite{defferrard2016convolutional,kipf2016semi}. 
DGCNN \cite{dgcnn} constructs \(k\)NN graphs for investigating correlated relationships among neighboring points with the EdgeConv operation, which encodes the graph local geometric information by propagating and aggregating edge features. 
3D-GCN \cite{3dgcn} proposes the deformable GCN kernel with learnable shapes and weights.
AdaptConv \cite{adaptconv} generates adaptive kernels for convolution on mutually correlated point pairs according to their edge features. 
More recently, \citet{deepgcns} transfers residual/dense connections and dilated convolutions to GCNs to train very deep GCNs, avoiding vanishing gradients. 
These works usually partition a point cloud into point sets by the \(k\)NN or radius ball query method, and design sophisticated local feature extractors for learning point sets features.

\begin{figure*}[htbp]
\centering
\begin{minipage}[b]{1.0\linewidth}
\begin{center}
\includegraphics[width=1.0\linewidth]{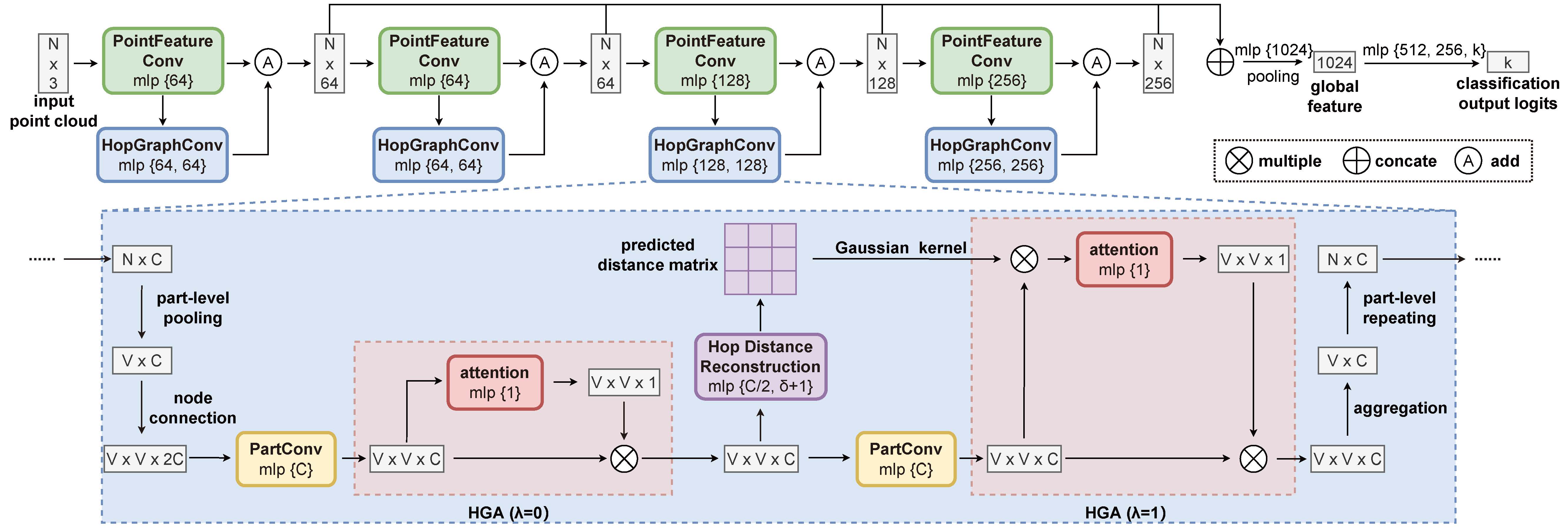}
\end{center}
\end{minipage}\\
\caption{
DHGCN architecture:
We feed the input point cloud to PointFeatureConv for extracting point-wise representations, which are then taken as input by Hop Graph Convolution (HopGraphConv), to extract more accurate local geometric representations.
Hop Graph Convolution:
The HopGraphConv layer takes the point features as input, and achieves part features through part-level pooling.
We construct a complete graph by taking parts as nodes and connecting each pair of them, and use PartConv and HGA to extract graph edge features.
We propose the self-supervised hop distance reconstruction task to predict the distance matrix of the complete graph from edge features.
\(\lambda\) controls whether the HGA embeds hop distance.
Finally, edge features are aggregated and repeated at the part-level, providing additional representations for the point-based backbone network.
}
\label{fig:netarchi}
\end{figure*}

\subsection{Attention Mechanism}
Attention mechanism exhibits the ability to extract relationships between representations by focusing on the most relevant parts of inputs to make decisions \cite{pointmlp}. 
\citet{vaswani2017attention} proposes Transformer, a model architecture relying solely on a self-attention (SA) mechanism, pioneering follow-up SA based works. 
Graph attention network (GAT) \cite{gat} computes the hidden representations of each graph node by attending to its neighbors' features with assigned attention weights. \citet{wang2019graphattenconv} proposes Graph attention convolution, implicitly assigning different attention importance to neighboring nodes and feature channels. 
\citet{guo2021pct} proposes PCT, which applies the offset-attention Transformer to learn the local context of the point cloud.

Unlike previous works that focus on designing sophisticated encoders for learning local features of grouped point sets, we propose the HGA to learn the contextual relationship between point sets, explicitly providing geometric information. This is driven by the fact that the inherent graph structure depicts the adjacent relationship between parts.

\section{Method}
\label{sec:method}
The overall architecture of the DHGCN is shown in Figure \ref{fig:netarchi}. 
DHGCN aims to learn the high-level contextual information from the pairwise neighboring relationship between point parts. 
We first voxelize the entire point cloud into parts and construct a graph by considering each part as a node. We define a distance matrix on the graph as the self-supervised signal, implying the degree of adjacency between nodes. 
Our DHGCN extracts parts features while also predicting the hop distance matrix, which is taken as input by the Hop Graph Attention (HGA) to embed the learned geometric information into point features.

\subsection{Volumetric Partition}
\label{sec:volumetricpartition}
This section discusses the details of point cloud volumetric partition and part-level graph construction. 
Given an input point cloud \(\mathcal{X}=\left\{x_{i} \mid i=\right.\) \(1,2, \ldots, N\} \in \mathbb{R}^{N \times 3}\) with \(N\) points, we voxelize it by mapping each point to a voxel part.
As a result, the volumetric partition splits \(\mathcal{X}\) into \(V={s}^3\) parts: \(\mathcal{P}=\left\{p_{i} \mid i=\right.\) \(1,2, \ldots, V\}\), where \(s\) is the split number, \(V\) is the number of parts, and \(p_i\) is a list containing the indices of points inside the \(i\)-th part.
For each part, we compute its up-scaled axis-aligned bounding box with the scale factor set to 1.2 for calculating the adjacency between parts in the geometric space. For parts with no points inside, we simply set its bounding box volume to 0. 

We can now construct a ground truth graph \(G=(\mathcal{V}, \mathcal{E})\) with each part serving as a graph node.
We define two nodes are connected by an edge if their bounding boxes are intersected.
We also introduce self-loops, connecting each node with itself.
The distance between two nodes is defined as the number of edges (i.e., hops) in their shortest path.
Therefore, \(G\)'s distance matrix \(D \in \mathbb{R}^{|\mathcal{V}| \times |\mathcal{V}|}\) implies the pairwise degree of adjacency between nodes quantitatively.
The maximum distance \(\delta\) is truncated to \(s+1\) hops and the self-loop distance is defined as 0 hop.

\subsection{Part Feature Extraction}
\label{sec:partfeatureextractor}
Point-based networks such as PointNet \cite{qi2017pointnet}, DGCNN \cite{dgcnn} and AdaptConv \cite{adaptconv} usually extract point-wise representations in each layer by using a point feature convolution (PointFeatureConv) function \(g_p\) with learnable parameters to map the input point cloud representations to a new set of \(C\)-dimensional point features \(\mathcal{H}=\left\{h_{i} \mid i=1,2, \ldots, N\right\} \in \mathbb{R}^{N \times C}\).
Then the point cloud's global features \(f_g \in \mathbb{R}^C\) are usually derived by applying the permutation-invariant max pooling to aggregate point-wise features globally:
\begin{equation}
    f_{g} = \max  _{j \in  \{1,2,\dots,N\} } h_j.
\end{equation}

In our DHGCN framework, we expect to achieve part-level features from the input point-wise features \(\mathcal{H}\). 
Similarly, we pool the point-wise features in \textit{part-level} to obtain part features \(\mathcal{F}=\left\{f_{i} \mid i=1,2, \ldots, V\right\} \in \mathbb{R}^{V \times C}\), where
\begin{equation}
    f_{i} = \max  _{j \in p_i} h_j
\end{equation}
and \(p_i\) contains the indices of points in the \(i\)-th part.

\subsection{Hop Graph Convolution Module}  
\label{sec:dynamichopconv}

\subsubsection{Part Convolution}
Given the input parts \(\mathcal{P}\) with its corresponding part features \(\mathcal{F}\), we initialize a complete graph  \(\tilde{G}=(\tilde{\mathcal{V}}, \tilde{\mathcal{E}})\) whose topology implies no specific contextual information, with parts serving as graph nodes for applying part-level graph convolution (PartConv).
Inspired by DGCNN \cite{dgcnn}, we define the edge feature between the \(i\)-th and \(j\)-th node as:
\begin{equation}
    e_{i j}=\left[f_{i}, f_{j}-f_{i}\right] \in \mathbb{R}^{2C},    
\end{equation}
where \([\cdot, \cdot]\) is the concatenation operation. 
\(e_{i j}\) explicitly combines shape information with neighborhood information encoded by the feature differences \cite{dgcnn}.
Our PartConv projects \(e_{ij}\) to a new set of edge features \(e_{ij}'\), attempting to extract more accurate local geometric representations \cite{adaptconv} as follows:
\begin{equation}
e_{i j}'=g_{m}\left(e_{ij}\right),
\end{equation}
where \(j \in \mathcal{N}(i) =\{j:(i, j) \in \tilde{\mathcal{E}}\}\), \(\mathcal{N}(i)\) is node indices connected with the \(i\)-th node. 
\(g_m(\cdot): \mathbb{R}^{C_{in}} \rightarrow \mathbb{R}^{C}\) is a MLP with learnable parameters.

\subsubsection{Hop Distance Reconstruction}
Since the low-level geometric distance can be represented by the degree of adjacent relationship between parts, we extract it by reconstructing the hop distance matrix of \(\tilde{G}\) with the proposed self-supervised task.
We denote the predicted hop distance of the \(i\)-th and \(j\)-th nodes of \(\tilde{G}\) as \(\tilde{D}_{i j}\) and consider predicting \(\tilde{D}_{i j}\) as a classification problem of \(\delta + 1\) categories (including self-loops), where \(\delta\) is the maximum distance.
In this sense, we predict \(\tilde{D}_{ij}\) by applying the MLP \(g_h\) to edge features \(e_{i j}'\):
\begin{equation}
    \tilde{D}_{i j} = g_h(e_{i j}'). 
\end{equation}

\begin{figure}[!h]
\centering
\begin{minipage}[b]{1.0\linewidth}
\begin{center}
\includegraphics[width=1.0\linewidth]{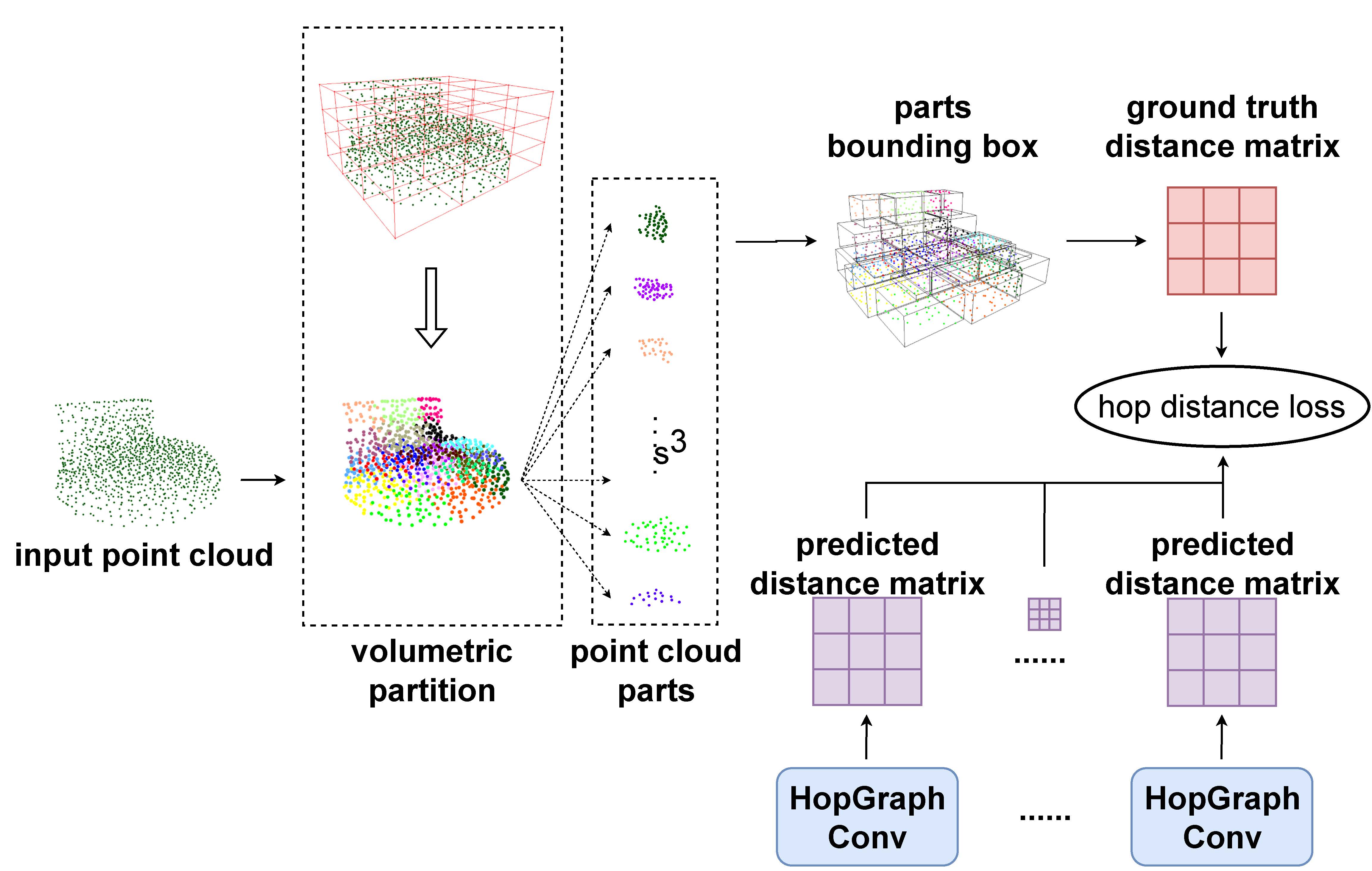}
\end{center}
\end{minipage}
\caption{
Given the input point cloud, we first voxelize it into parts.
For each part, we compute its scaled axis-aligned bounding box to calculate the adjacent relation between parts. We construct a ground truth graph along with its distance matrix for supervision in each layer.
}
\label{fig:hop_distance}
\end{figure}

As shown in Figure \ref{fig:hop_distance}, the predicted distance matrix \(\tilde{D}\) is updated in each layer supervised by the ground truth distance matrix \(D\).
Concretely, we propose the hop distance loss \(\mathcal{L}_h\) to measure the discrepancy between \(\tilde{D}\) and \(D\) in each layer:
\begin{equation}
\mathcal{L}_h = \sum_{i}^{V} \sum_{j}^{V} \operatorname{CE}(\tilde{D}_{i j}, D_{i j}),
\end{equation}
where \(\operatorname{CE}\) is a cross-entropy function.

\subsubsection{Hop Graph Attention}
The Hop Graph Attention aims to embed the learned geometric structure information into high-level point cloud contextual features by assigning more attention weights to edge features between neighboring parts in the geometric space (i.e., parts with low hops).
Therefore, we compute the hop attention coefficient of \(e_{i j}'\) as follows:
\begin{equation}
    t_{i j} = g_a\left( \lambda \cdot \mathbb{G}(\tilde{D}_{ij}) e_{ij}' + (1-\lambda)  e_{ij}' \right),
\end{equation}
where \(j \in \mathcal{N}(i)\), and \(g_a : \mathbb{R}^{C} \rightarrow \mathbb{R}\) is a shared attention MLP.  \(\mathbb{G}(x) = \frac{1}{\sqrt{2 \pi}} \exp \left(-\frac{1}{2 \sigma ^2} x^2\right)\) represents Gaussian kernel.
\(\lambda \in  \{0,1\}\) is a switch factor that controls whether to embed the hop distance. When \(\lambda=0\), the HGA equals to general SA to extract preliminary features. 

The hop attention coefficient \(t_{i j}\) indicates the learned importance of part \(p_j\) to \(p_i\), which is inferred from the predicted hop distance \(\tilde{D}_{ij}\).
Since the hop distance depicts the adjacent relationship between parts, explicitly multiplying \(\tilde{D}_{ij}\) by the edge features \(e_{ij}'\) embeds the low-level geometric distance into the high-level contextual features, producing more expressive attention weights.

We normalize coefficients using softmax function to make it comparable across all connected neighbors as follows:
\begin{equation}
    \alpha_{i j}=\operatorname{softmax}_j\left(t_{i j}\right)=\frac{\exp \left(t_{i j}\right)}{\sum_{k \in \mathcal{N}(i)} \exp \left(t_{i k}\right)}
\end{equation}

We can now calculate the learned part features \(\tilde{f}_i\) by aggregating edge features between the \(i\)-th node and all other connected nodes distinctively:
\begin{equation}
\tilde{f}_i = \max  _{j \in \mathcal{N}(i)} \alpha_{ij} \cdot e'_{ij}.
\end{equation}

\subsubsection{Revising Point-wise Features}

For each part-level feature \(\tilde{f}_i\), we repeat it \(|p_i|\) times to align it with corresponding points, constructing the revised point-wise features \( \tilde{\mathcal{H}} = \{\tilde{h}_i \mid i=1,2,...,N \}\) by fusing with the original point-wise features \(\mathcal{H}\) through element-wise addition for providing more expressive representations for our point-based backbone network: 
\begin{equation}
    \tilde{h}_j = \tilde{f}_i + h_j , j \in p_i.
\end{equation}

\begin{figure*}[htbp]
\begin{center}
\begin{minipage}[b]{0.8\linewidth}
\begin{center}
\begin{minipage}[b]{0.15\linewidth}
\begin{center}
\raisebox{0ex}{\footnotesize{attention maps}}
\raisebox{3.5ex}{\footnotesize{(Ours)}}
\end{center}
\end{minipage}
\begin{minipage}[b]{0.12\linewidth}
\begin{center}
\includegraphics[width=1.0\linewidth]{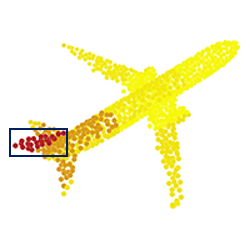}
\end{center}
\end{minipage}
\begin{minipage}[b]{0.12\linewidth}
\begin{center}
\includegraphics[width=1.0\linewidth]{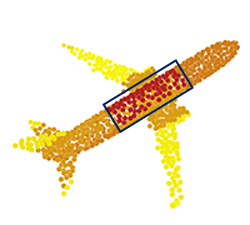}
\end{center}
\end{minipage}
\begin{minipage}[b]{0.12\linewidth}
\begin{center}
\includegraphics[width=1.0\linewidth]{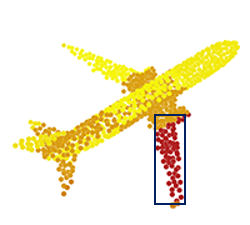}
\end{center}
\end{minipage}
\begin{minipage}[b]{0.12\linewidth}
\begin{center}
\includegraphics[width=1.0\linewidth]{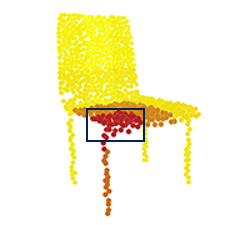}
\end{center}
\end{minipage}
\begin{minipage}[b]{0.12\linewidth}
\begin{center}
\includegraphics[width=1.0\linewidth]{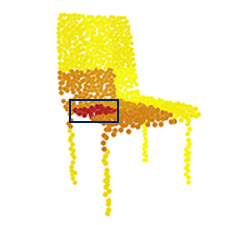}
\end{center}
\end{minipage}
\begin{minipage}[b]{0.12\linewidth}
\begin{center}
\includegraphics[width=1.0\linewidth]{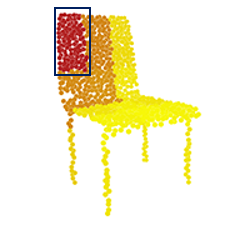}
\end{center}
\end{minipage}\\
\begin{minipage}[b]{0.15\linewidth}
\begin{center}
\raisebox{0ex}{\footnotesize{feature distance}}
\raisebox{3.5ex}{\footnotesize{(Ours)}}
\end{center}
\end{minipage}
\begin{minipage}[b]{0.12\linewidth}
\begin{center}
\includegraphics[width=1.0\linewidth]{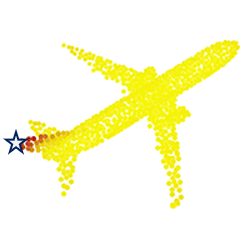}
\end{center}
\end{minipage}
\begin{minipage}[b]{0.12\linewidth}
\begin{center}
\includegraphics[width=1.0\linewidth]{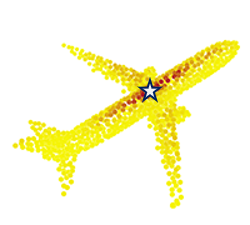}
\end{center}
\end{minipage}
\begin{minipage}[b]{0.12\linewidth}
\begin{center}
\includegraphics[width=1.0\linewidth]{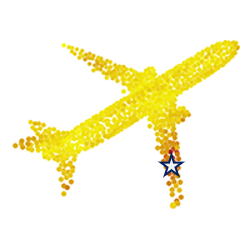}
\end{center}
\end{minipage}
\begin{minipage}[b]{0.12\linewidth}
\begin{center}
\includegraphics[width=1.0\linewidth]{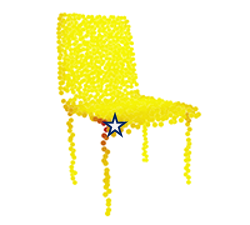}
\end{center}
\end{minipage}
\begin{minipage}[b]{0.12\linewidth}
\begin{center}
\includegraphics[width=1.0\linewidth]{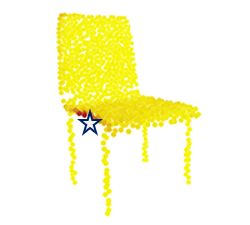}
\end{center}
\end{minipage}
\begin{minipage}[b]{0.12\linewidth}
\begin{center}
\includegraphics[width=1.0\linewidth]{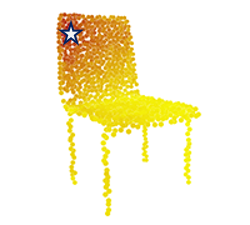}
\end{center}
\end{minipage}\\
\begin{minipage}[b]{0.15\linewidth}
\begin{center}
\raisebox{0ex}{\footnotesize{feature distance}}
\raisebox{3.5ex}{\footnotesize{(DGCNN)}}
\end{center}
\end{minipage}
\begin{minipage}[b]{0.12\linewidth}
\begin{center}
\includegraphics[width=1.0\linewidth]{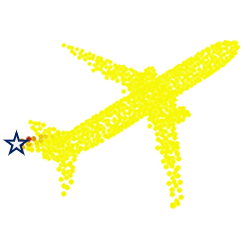}
\end{center}
\end{minipage}
\begin{minipage}[b]{0.12\linewidth}
\begin{center}
\includegraphics[width=1.0\linewidth]{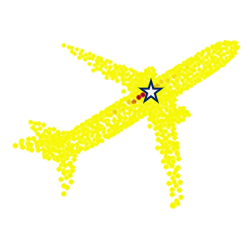}
\end{center}
\end{minipage}
\begin{minipage}[b]{0.12\linewidth}
\begin{center}
\includegraphics[width=1.0\linewidth]{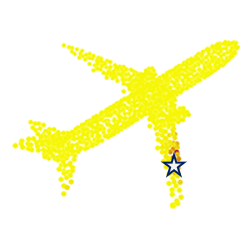}
\end{center}
\end{minipage}
\begin{minipage}[b]{0.12\linewidth}
\begin{center}
\includegraphics[width=1.0\linewidth]{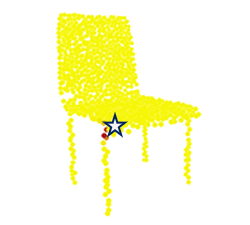}
\end{center}
\end{minipage}
\begin{minipage}[b]{0.12\linewidth}
\begin{center}
\includegraphics[width=1.0\linewidth]{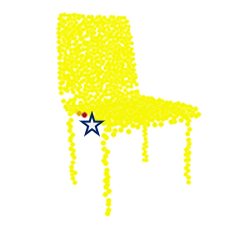}
\end{center}
\end{minipage}
\begin{minipage}[b]{0.12\linewidth}
\begin{center}
\includegraphics[width=1.0\linewidth]{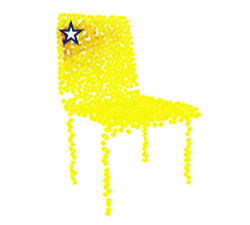}
\end{center}
\end{minipage}
\end{center}
\end{minipage}
\begin{minipage}[b]{0.04\linewidth}
\begin{center}
\includegraphics[width=1.6\linewidth]{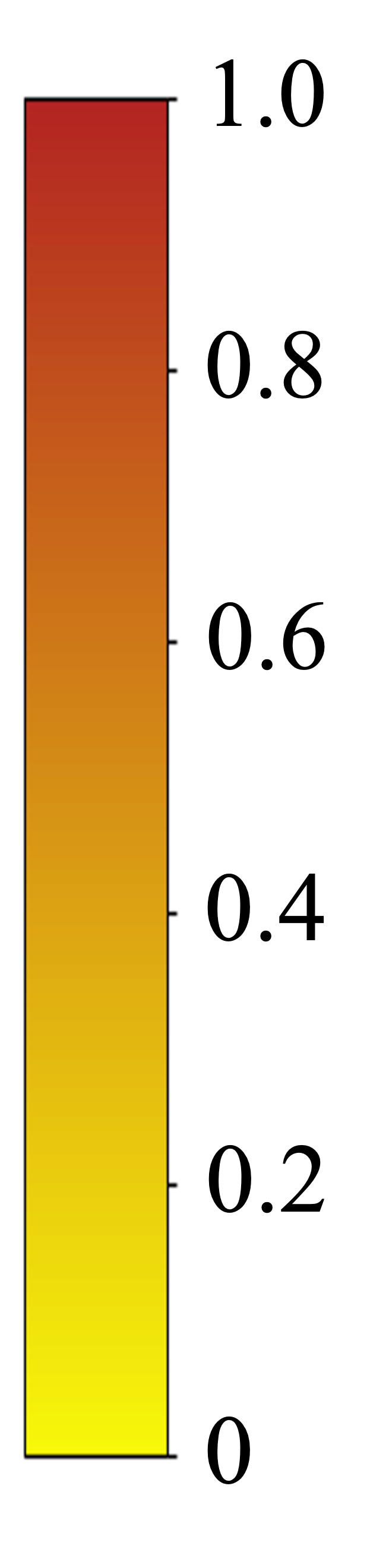}
\end{center}
\end{minipage}
\end{center}
\caption{Attention maps (row 1) for different query parts and feature distance from the query point (indicated by star, row 2 (ours) and row 3 (DGCNN)) to all other points, with yellow to red indicating increasing attention weight or closer distance.}
\label{fig:feature_vis}
\end{figure*}

Figure \ref{fig:feature_vis} visualizes our attention maps and the comparison of feature distance between our method and the DGCNN backbone.
The attention maps (row 1) depict that the predicted hop distance is consistent with the geometric relationship of each part, which enables our learned features \(\tilde{\mathcal{H}}\) (row 2) to be more distinctive than DGCNN (row 3), and more related to both geometric distance and shape information.

\subsubsection{Dynamic Hop Distance}
Previous point learning works \cite{dgcnn, pointnet++, guo2021pct} tend to construct a local graph for each selected center point through \(k\)NN in feature space, and update center points and corresponding graphs dynamically in each layer.
Since their edge features contribute equally in aggregation, these methods can be viewed as learning local features from local point sets using a GCN with a local receptive field.

Different from these works, we construct a complete global graph \(\tilde{G}\) rather than a local graph, allowing every node to attend to every other node, making the receptive field cover the entire point cloud globally.
With the aid of the distance reconstruction task,
\(\tilde{G}\)'s hop distance matrix is dynamically updated in each layer, which is enabled by the Gaussian kernel for providing more attention to neighboring parts and less attention to distant parts, allowing edge features to contribute distinctively in aggregation.

We calculate the proposed hop distance loss \(\mathcal{L}_h^{(l)}\) in each layer to provide strong supervision for updating the distance matrix. Finally, 
the self-supervised loss of our DHGCN for the hop distance prediction task is defined as:
\begin{equation}
    \mathcal{L} = \sum_{l}^{L}\mathcal{L}_h^{(l)},
\end{equation}
where \(L\) denotes the number of HopGraphConv layers.

\section{Experimental Results}
\label{sec:results}
\subsection{Experimental Setting}
We implement our method in PyTorch. 
The SGD optimizer is used for all experiments.
We use one TITAN RTX GPU for training. 
The training batch size is set to $32$. 
Following DGCNN, we set the dropout rate to $0.5$ and the random seed to $1$.
The learning rate is set to $0.1$ under the cosine learning scheduler with the $0.9$ momentum and the 0.0005 weight decay. 
For classification, we set the split number $s=3$ by default, i.e., the point cloud is split into $3^3=27$ parts, and the max distance $\delta=4$, i.e., the hop distance will be classified into $5$ categories (including hop = $0$), and we set $s=5$ for segmentation tasks for a more fine-grained splitting. 
We use $4$ heads for multi-head attention in the Hop Graph Attention module. \(\sigma\) in Gaussian kernel is set to $1$ by default.

\begin{table*}
\begin{center}
\begin{tabular}{l c c c || l c c c}
\hline
Methods & \tabincell{c}{Pretrained\\dataset} & \# Points & Acc. & Methods & \tabincell{c}{Pretrained\\dataset} & \# Points & Acc. \\
\hline
LatentGAN \shortcite{latentgan}  & SN & 2k & 85.7 & FoldingNet \shortcite{yang2018foldingnet} & MN & 2k & 84.4\\
FoldingNet \shortcite{yang2018foldingnet} & SN & 2k & 88.4 & LatentGAN  \shortcite{latentgan} & MN & 2k & 87.3\\
PointCapsNet \shortcite{zhao20193d}  & SN & 2k & 88.9 & PointCapsNet \shortcite{zhao20193d} & MN & 1k & 87.5\\
VIPGAN \shortcite{han2019view}  & SN & 2k & 90.2 & Multi-task \shortcite{hassani2019unsupervised}  & MN & 2k & 89.1\\
STRL \shortcite{huang2021spatio}  & SN & 2k & 90.9 & MAP-VAE \shortcite{han2019mapvae}  & MN & 2k & 90.2\\
SSC (RSCNN) \shortcite{chen2021shape} & SN  & 2k & 92.4 & GraphTER \shortcite{gao2020graphter} & MN & 1k & 92.0\\
CrossPoint \shortcite{afham2022crosspoint} & SN & 2k & 91.2 & GLR (RSCNN) \shortcite{rao2020global} & MN & 1k & 92.2\\
\hdashline
\textbf{DHGCN (DGCNN}) & SN & 2k & \textbf{93.2} &  DHGCN (DGCNN) & MN & 1k & 93.0\\
\bf{DHGCN (AdaptConv)} & SN & 2k & \textbf{93.2} & \bf{DHGCN (AdaptConv)} & MN & 1k & \bf{93.3}\\
\hline
\end{tabular}
\end{center}
\caption{Classification results of \textit{unsupervised} methods (including ours) on ModelNet40. `SN/MN' denotes `ShapeNet/ModelNet40' and `\# Points' indicates the point number in pretraining. }
\label{table:unsuperMN}
\end{table*}

\subsection{Pretraining}
\textbf{Pretrained datasets.}
Our DHGCN uses ShapeNet \cite{chang2015shapenet} for pretraining. The dataset has $57,448$ models with $55$ categories, and all models will be used for our self-supervised pretraining task. Following previous work, we use $2,048$ points as input. Note that some methods are pretrained on ModelNet40 \cite{wu20153d}. We also follow this setting, and use the train set ($9,840$ training models) of ModelNet40 for pretraining. 
As for ModelNet40, we use $1,024$ points as input which is the same with GraphTER.

\textbf{Training details.}
The DHGCN is a part-level plug-and-play module that is compatible with point-based backbone networks, by implementing the PointFeatureConv with backbone modules. 
We use several different point-based backbone networks under the linear protocol of unsupervised representation learning to verify the effectiveness of our self-supervised reconstruction task.
This training strategy is a two-stage paradigm. The first stage is to pretrain the network with only the proposed self-supervised task, and the second stage is to \textit{freeze} the pretrained model and train a linear classifier only for downstream tasks, i.e., 3D object classification and shape part segmentation. 

\subsection{Downstream Tasks}

\textbf{3D object classification on ModelNet40.}
We use ModelNet40 for 3D point cloud classification. 
Data split protocols following PointNet. 
The dataset contains $9,840$ models for training and $2,468$ models for testing, involving a total of $40$ categories. 
The sampling strategy in PointNet is adopted to sample each point cloud into $1,024$ points. 
We only use normalized coordinates as input without considering normals. 

Table \ref{table:unsuperMN} shows the comparison results of our method and the SOTA unsupervised methods. 
We divide the results based on two pretrained datasets. 
As we can see, our DHGCN notably outperforms recent works, achieving SOTA performance on both pretrained datasets ($93.2\%$ for ShapeNet and $93.3\%$ for ModelNet40), exceeding SSC by $0.8\%$ with pretraining on ShapeNet and surpassing GraphTER by $1.3\%$ with pretraining on ModelNet40.

\setlength{\tabcolsep}{4pt}
\begin{table}[tbp]
\begin{center}
\begin{tabular}{l c c c c c}
\hline
\multirow{2}*{Methods} & \multicolumn{5}{c}{Limited training data ratios}\\
\cline{2-6}
~ & 1\% & 2\% & 5\% & 10\% & 20\%\\
\hline
FoldingNet \shortcite{yang2018foldingnet} & 56.4 & 66.9 & 75.6 & 81.2 & 83.6\\
MAE3D \shortcite{jiang2023masked} & 61.7 & 69.2 & 80.8 & 84.7 & 88.3\\
\hdashline
\textbf{DHGCN} & \textbf{62.7} & \textbf{72.2} & \textbf{81.3} & \textbf{86.1} & \textbf{89.1}\\
\hline
\end{tabular}
\end{center}
\caption{Comparison results of 3D object classification with limited training data (different ratios) on ModelNet40. DGCNN is taken as the backbone.
}
\label{table:limited_data}
\end{table}

\textbf{Classification with limited data}.
Following previous works \cite{yang2018foldingnet}, we further train a linear classifier with only limited sampled training data of ModelNet40 to evaluate the pretrained model's generalizability on all test data. 
As shown in Table \ref{table:limited_data}, our method with DGCNN as backbone can achieve $89.1\%$ with $20\%$ data used for training and $86.1\%$ with $10\%$ data.
In the extreme case of using only $1\%$ data, our DHGCN achieves $62.7\%$ accuracy, exceeding that of MAE3D by $1\%$ and FoldingNet by $6.3\%$.
Our DHGCN achieves SOTA results in all $5$ training data ratios, demonstrating that the  features learned by our self-supervised task can be easily generalized to the point cloud classification task even with limited training data.

\textbf{Classification on real-world dataset ScanObjectNN.}
We also conduct the classification experiment on the real-world scanning dataset ScanObjectNN \cite{uy2019revisiting}, which poses great challenges for point cloud classification methods due to the involved cluttered background, noisy perturbations and occluded incomplete data. 
This dataset contains $15$ categories, totally $2,902$ unique object instances. 
Here we follow the official data split strategy on three dataset variants: OBJ\_ONLY, OBJ\_BG and PB\_T50\_RS, and conduct pertaining on ShapeNet.

As shown in Table \ref{table:scanobjectnn}, DHGCN achieves the best results of \(85.9\%\) on OBJ\_BG variant, exceeding all compared SOTA unsupervised methods by at least $4.2\%$.
Our DHGCN even achieves comparable results with supervised methods, e.g., surpassing the DGCNN backbone by $3.1\%$ on OBJ\_BG.
As for the PB\_T50\_RS variant, our method achieves an accuracy of \(81.9\%\), slightly lower than the other two variants. 
We suspect that the perturbation noise in PB\_T50\_RS disturbs the adjacency relationship between parts (i.e., some points are incorrectly split into adjacent parts during voxelization), thus degrading the power of the learned hop distance in depicting point cloud intrinsic geometric structure.

Furthermore, despite being pretrained on the ShapeNet dataset (synthetic data), our downstream classification results on the real-world ScanObjectNN reveal that the learned geometric information is useful in mitigating the domain gap between synthetic and real-world data.

\setlength{\tabcolsep}{3pt}
\begin{table}
\begin{center}\fontsize{9}{10} \selectfont
\begin{tabular}{l c c c c}
\hline
Methods & Sup. & OBJ\_ONLY & OBJ\_BG & PB\_T50\_RS\\
\hline
PointNet \shortcite{qi2017pointnet}      & \cmark & 79.2 & 73.3 & 68.2\\
PointNet++ \shortcite{pointnet++} & \cmark & 84.3 & 82.3 & 77.9\\
PointCNN  \shortcite{li2018pointcnn}  & \cmark & 85.5 & 86.1 & 78.5\\
DGCNN  \shortcite{dgcnn}      & \cmark & 86.2 & 82.8 & 78.1\\
Point-BERT \shortcite{yu2022point}  & \cmark & 88.1 & 87.4 & 83.1\\
Point-MAE \shortcite{pang2022masked}  & \cmark & 88.3 & 90.0 & 85.2\\
\hline
Jigsaw \shortcite{sauder2019self} & \xmark & - & 59.5 & - \\
OcCo \shortcite{wang2021unsupervised} & \xmark & - & 78.3 & - \\
STRL \shortcite{huang2021spatio} & \xmark & - & 77.9 & - \\
CrossPoint \shortcite{afham2022crosspoint} & \xmark & - & 81.7 & - \\
\hdashline
\bf{DHGCN}          & \xmark & \textbf{85.0} & \textbf{85.9} &  \textbf{81.9}\\
\hline
\end{tabular}
\end{center}
\caption{Classification results of our method and state-of-the-art methods on ScanObjectNN. DGCNN is used as backbone. `Sup.' denotes the method is supervised (\cmark) or unsupervised (\xmark). Results of Jigsaw, OcCo, and STRL are from CrossPoint, and ``-'' indicates no previous results.
}
\label{table:scanobjectnn}
\end{table}

\textbf{Part segmentation on ShapeNet Part.}
We evaluate DHGCN for the shape part segmentation task on ShapeNet Part dataset \cite{yi2016scalable}, which contains $16,881$ models from $16$ categories. Each model involves $2$ to $6$ parts, with a total number of $50$ distinct part labels. Following PointNet, we sample or interpolate each model to $2,048$ points and only use point coordinates as input. 

We use mean Intersection-over-Union (mIoU) as the evaluation metric, and two types of mIoU are reported in Table \ref{table:partseg}. 
Our self-supervised method achieves SOTA performance, which exceeds all recent unsupervised methods.

\setlength{\tabcolsep}{3pt}
\begin{table}[htb]
\begin{center}
    \begin{tabular}{l c c c}
        \hline
         Methods & Sup. & Class mIOU & Instance mIOU\\
         \hline
PointNet \shortcite{qi2017pointnet} & \cmark & 80.4 & 83.7\\
PointNet++ \shortcite{pointnet++} & \cmark & 81.9 & 85.1\\
DGCNN \shortcite{dgcnn} & \cmark & 82.3 & 85.2\\
KPConv \shortcite{thomas2019kpconv} & \cmark & 85.1 & 86.4\\
PAConv \shortcite{xu2021paconv}& \cmark  & 84.2 & 86.0\\
Point-BERT \shortcite{yu2022point} & \cmark & 84.1 & 85.6\\
    \hline
LatentGAN \shortcite{latentgan} & \xmark & 57.0 & -\\
MAP-VAE \shortcite{han2019mapvae} & \xmark & 68.0 & - \\
GrpahTER  \shortcite{gao2020graphter} & \xmark & 78.1 & 81.9\\
CTNet \shortcite{jiang2023unsupervised} & \xmark & 75.5 & 79.2\\
\hdashline
\bf{DHGCN} & \xmark & \textbf{82.9} & \textbf{84.9}\\
\hline
    \end{tabular}
\end{center}
\caption{Shape part segmentation results of our method and state-of-the-art techniques on ShapeNet Part dataset. PAConv is used as backbone. `Sup.' denotes the method is supervised learning (\cmark) or unsupervised learning (\xmark).}
\label{table:partseg}
\end{table}

\textbf{Part segmentation with limited data.}
We \textit{freeze} the pretrained model and randomly sample $1\%$ and $5\%$ of the train set of ShapeNet Part to train several MLPs for evaluating the segmentation task with the unsupervised paradigm.
Results shown in Table \ref{table:limited_data_seg} demonstrate that our DHGCN using PAConv as backbone achieves SOTA performance, i.e., $76.9\%$ with $1\%$ training data and $81.9\%$ with $5\%$ training data for instance mIoU, which exceeds recent unsupervised methods.
These results reveal that the features learned by our self-supervised hop distance reconstruction task are more expressive than other unsupervised methods in terms of part segmentation, under extremely limited training data.

\begin{table}
\begin{center}
\begin{tabular}{p{3.4cm} p{1.8cm}<{\centering} p{1.8cm}<{\centering}}
\hline
\multirow{2}*{Methods} & \multicolumn{2}{c}{Limited training data ratios}\\
\cline{2-3}
~ & 1\% & 5\%\\
\hline
SO-Net  \shortcite{li2018so} & 64.0 & 69.0\\
PointCapsNet \shortcite{zhao20193d} & 67.0 & 70.0\\
Multi-task \shortcite{hassani2019unsupervised} & 68.2 & 77.7\\
PointContrast \shortcite{xie2020pointcontrast}  & 74.0 & 79.9\\
SSC (RSCNN) \shortcite{chen2021shape} & 74.1 & 80.1\\
\hdashline
\textbf{DHGCN} & \textbf{76.9} & \textbf{81.9}\\
\hline
\end{tabular}
\end{center}
\caption{Comparison results of shape part segmentation with limited training data (different ratios) on ShapeNet Part. PAConv is taken as the backbone.
}
\label{table:limited_data_seg}
\end{table}

\textit{Please refer to supplementary for more results.}

\subsection{Ablation Studies}
\textbf{Attention mechanism.}
We conduct an ablation study for several model settings to verify the HGA's effectiveness, which embeds the learned hop distance matrix into edge weights.
We denote the SA option as our baseline, whose switch factor \(\lambda\) is set to $0$ in both HGA layers. 
HGA will degrade to general SA in this case, and the hop distance loss is disabled. Note that we train this baseline in a \textit{supervised} manner.
We compare the effect of whether the hop distance loss is calculated in each layer or only the last layer.
Accuracy results of point cloud classification and hop distance prediction are reported in Table \ref{table:attention}. 

With the aid of HGA, the classification accuracy significantly exceeds that of SA baseline by $0.5\%$. 
In addition, calculating the hop distance loss in each layer leads to both higher hop distance prediction accuracy (\(94.6\% \) versus \(93.3\%\)) and point cloud classification accuracy ($93.3\%$ versus $93.1\%$). 
The strong supervision of our hop distance loss leads to a more accurate learned hop distance matrix, thus producing better performance. 

\setlength{\tabcolsep}{4pt}
\begin{table}
\begin{center}
\begin{tabular}{l c l c c}
\hline
Attention & Sup. & Loss & Distance Acc. & Acc.\\
\hline
SA & \cmark & $\mathcal{L} = \mathcal{L}_c $ & - & 92.8\\
HGA & \xmark & $\mathcal{L} = \mathcal{L}_h^{(-1)}$ & 93.3 & 93.1\\
HGA & \xmark & $\mathcal{L} = \sum_{l}^{L}\mathcal{L}_h^{(l)}$ & \bf{94.6} & \bf{93.3}\\
\hline
\end{tabular}
\end{center}
\caption{
Different attention mechanisms. Experiments are conducted on ModelNet40 with AdaptConv as the backbone.
SA denotes self-attention, and HGA denotes Hop Graph Attention.
Accuracy results of hop distance prediction and point cloud classification are reported.
}
\label{table:attention}
\end{table}

\textbf{Gaussian kernel.}
The predicted hop distance will be processed by the Gaussian kernel \(\mathbb{G}\) to provide more attention to neighboring parts (i.e., smaller distances yield higher weights and vice versa). 
The parameter \(\sigma^2\) of \(\mathbb{G}\) controls the decay rate between distance and edge weight. 
A small \(\sigma^2\) causes the edge weights between remote parts to decay rapidly. 
For example, when \(\sigma^2=0.2\), the weights of parts over \(1\)-hop distance converge to $0$. At this point, the receptive field degrades to $1$-hop (i.e., local neighbors), resulting in a lower accuracy of $92.6\%$, as shown in Table \ref{table:ablation}. 
On the contrary, with a larger $\sigma^2$, the weights decay gently as the distance increases, enabling nodes to contribute more equally. 
However, this leads to a reduction in distinction due to distance, achieving only $93.0\%$ ($\sigma^2=2.0$) and $92.9\%$ ($\sigma^2=5.0$). 
The model achieves the highest accuracy of $93.3\%$ when $\sigma^2=1.0$.

\begin{table}
\begin{center}
\begin{tabular}{c| c c c c c}
    \hline
     \multicolumn{6}{c}{Gaussian kernel}\\
     \hline
     $\sigma^2$ & 0.2 & 0.5 & \textbf{1.0} & 2.0 & 5.0\\
     Acc. & 92.6 & 93.2 & \textbf{93.3} & 93.0 & 92.9 \\
     \hline
\end{tabular}
\end{center}
\caption{Ablation results on different \(\sigma^2\) in the Gaussian kernel. Experiments are conducted on ModelNet40 for classification with AdaptConv as the backbone.}
\label{table:ablation}
\end{table}

\section{Conclusion}
This paper proposes a novel self-supervised part-level hop distance reconstruction task and a novel hop distance loss to learn  contextual relationships between point parts. 
The dynamically updated hop distances are embedded as attention weights by the proposed HGA for determining point parts' importance in feature aggregation. 
Our DHGCN 
can be easily incorporated into point-based backbones. 
We 
outperform SOTA unsupervised methods on both downstream classification and part segmentation tasks. 
Our model is less effective for data with large perturbations as noise leads to less accurate splitting of parts, which tends to produce misleading adjacent relationships. This will be explored in future. 

\section*{Acknowledgments}
This work is supported in part by the National Key Research and Development Program of China (Grant Number 2022ZD04014) and the Shaanxi Province Key Research and Development Program (Grant Number 2022QFY11-03).

\bibliography{aaai24}

\end{document}